\documentclass[conference]{IEEEtran}
\IEEEoverridecommandlockouts
\usepackage{cite}
\usepackage{amsmath,amssymb,amsfonts}
\usepackage{algorithmic}
\usepackage{graphicx}
\usepackage{textcomp}
\usepackage{xcolor}
\usepackage{booktabs}
\usepackage{multirow}
\usepackage[hyphens]{url}
\def\BibTeX{{\rm B\kern-.05em{\sc i\kern-.025em b}\kern-.08em
    T\kern-.1667em\lower.7ex\hbox{E}\kern-.125emX}}

\newcommand\blfootnote[1]{%
  \begingroup
  \renewcommand\thefootnote{}\footnote{#1}%
  \addtocounter{footnote}{-1}%
  \endgroup
}

\begin{document}

\title{Pattern-Aware Graph Neural Networks for Handling Missing Data}


\author{\IEEEauthorblockN{Minett Tran, Taehee Jeong}
\IEEEauthorblockA{\textit{San Jose State University} \\
minett.tran, taehee.jeong @sjsu.edu}
{\thanks{This work was supported in part by a Mobilint Grant awarded to San Jose State University. (Corresponding author: Taehee Jeong)}
}
}

\maketitle

\begin{abstract}
Missing data is ubiquitous in real-world datasets. Traditional methods either discard incomplete samples or apply imputation techniques that ignore potentially informative missingness patterns, implicitly assuming that missingness occurs randomly. However, missingness patterns might provide additional information. We propose pattern-aware graph neural networks that explicitly encode which features are missing alongside observed values. We used four encoding strategies---learned embeddings, frozen random embeddings, statistical features, and hierarchical representations---across seven UCI datasets with naturally occurring missingness. Our Pattern-aware methods achieve substantial improvements over baselines, with an average improvement of 17\% in balanced accuracy and 22\% in F1-macro across all datasets. The benefits vary significantly by dataset: annealing shows dramatic improvement (+80\% balanced accuracy), while hepatitis and soybean show minimal gains (+4--5\%). Notably, even simple random pattern embeddings perform comparably to learned embeddings (0.650 vs 0.663 balanced accuracy), suggesting that distinguishing between patterns may be more important than task-specific optimization. Our ablation study reveals that attention mechanisms, while helpful, are not critical when pattern information is available---simple mean aggregation with pattern awareness achieves 0.640 balanced accuracy compared to 0.645 for attention-based variants.
Our code and data are available at \url{https://github.com/TranMinett/pattern_aware_GRAPE}.
\end{abstract}

\begin{IEEEkeywords}
Graph Neural Networks, Missing Data, Tabular Data, Pattern Encoding, Bipartite Graphs
\end{IEEEkeywords}

\section{Introduction}
\renewcommand{\footnoterule}{%
  \kern -3pt
  \hrule width 3in height 1pt
  \kern 2pt
}
\blfootnote{2026 International Conference on Advances in Artificial Intelligence and Machine Learning (AAIML), 20-22 March 2026, IEEE Copyright 2026}

Missing data pervades real-world datasets: medical records often contain missing diagnostic tests, survey respondents frequently skip questions, and sensors periodically fail. The traditional approach handles such incompleteness either by deleting samples with missing values or by imputing them using methods such as mean substitution or k-nearest neighbors. However, these approaches rest on a critical assumption---that missingness occurs randomly.

Consider a clinical setting where physicians order extensive diagnostic tests only for patients presenting severe symptoms. In this scenario, which tests are missing reveals information about disease severity independent of the test results themselves. Traditional imputation methods 
ignore this signal, focusing solely on estimating the unobserved values rather than leveraging the informative structure of missingness.

Graph Neural Networks (GNNs) offer an alternative paradigm for handling incomplete data. GRAPE \cite{you2020handling} represents each sample data as a bipartite graph where feature nodes connect only to observed values, naturally accommodating arbitrary missingness patterns without imputation. However, GRAPE treats all missingness patterns identically: two patients with different patterns of missing tests receive equivalent graph structures provided they share the same values for observed features, discarding potentially valuable information about which measurements were deemed necessary.

This raises a fundamental question: does explicitly encoding missingness patterns improve predictive performance? And critically, under what conditions does this added complexity justify itself? These questions carry practical significance beyond academic interest, as pattern-aware methods introduce architectural complexity and computational overhead that may not always be warranted.

We present pattern-aware extensions to GRAPE that explicitly encode missingness patterns through various embedding strategies. We developed four approaches: learned embeddings that train end-to-end with the model, random frozen embeddings that never update during training, statistical features derived from missingness patterns, and hierarchical encoding that learns compressed pattern representations. Through evaluation across seven diverse datasets, we observed substantial but heterogeneous improvements. 

Our main contributions are:

\begin{itemize}
\item Pattern-aware graph neural networks that incorporate missingness patterns, achieving 17\% average improvement in balanced accuracy across seven datasets
\item Evidence that simple random embeddings (0.650 balanced accuracy) perform comparably to learned embeddings (0.663), suggesting pattern distinction matters more than optimization
\item Demonstration that attention mechanisms become less critical with pattern information---simple mean aggregation achieves competitive results (0.640 vs 0.645)
\item Analysis revealing dataset-specific effectiveness ranging from 80\% improvement (annealing) to 4\% (hepatitis, soybean)
\end{itemize}

\begin{figure*}[t]
\centering
\includegraphics[width=\textwidth]{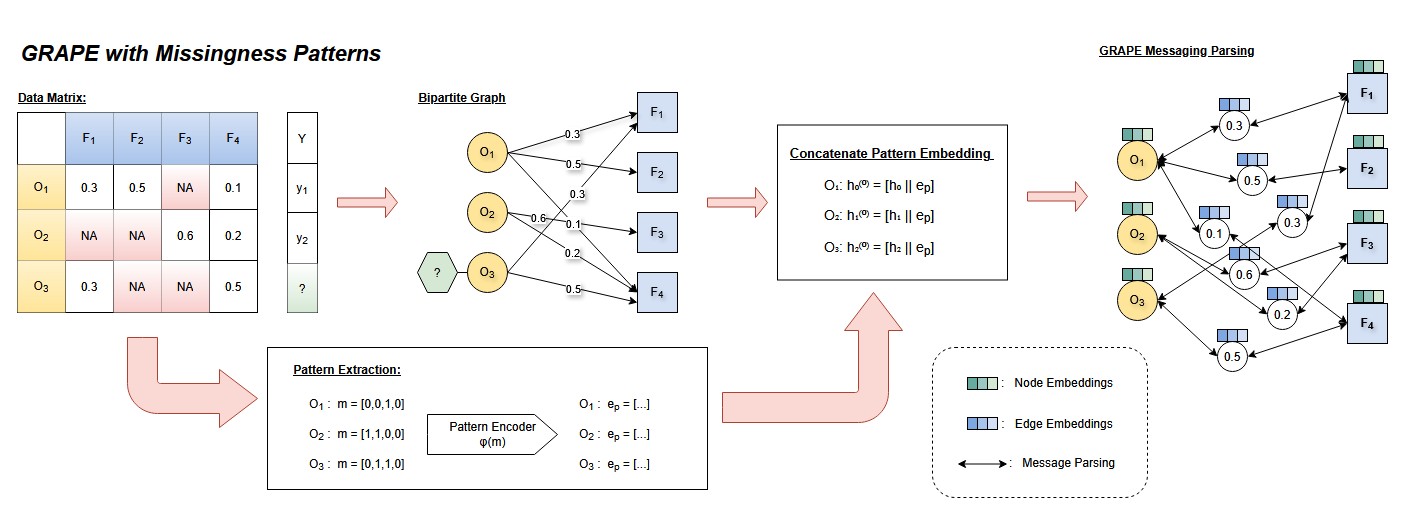}
\caption{Pattern-aware GRAPE architecture. Left: A data matrix with missing values (NA) is converted to bipartite graphs where edges exist only for observed features. Center: Binary missingness patterns $\mathbf{m}$ are extracted and encoded via $\phi(\mathbf{m})$ to produce pattern embeddings $\mathbf{e}_p$. Right: Pattern embeddings are concatenated with observation node representations before message passing, enabling the model to distinguish samples with different missingness patterns.}
\label{fig:architecture}
\end{figure*}

\section{Background}
\label{sec:background}

\subsection{Missing Data Mechanisms}

The missing data literature distinguishes three fundamental mechanisms \cite{little2019statistical,rubin1976inference}. Missing Completely At Random (MCAR) occurs when missingness is independent of both observed and unobserved data---for example, equipment randomly failing regardless of patient characteristics. Missing At Random (MAR) occurs when missingness depends on observed data but not on the missing values themselves---for instance, younger patients being less likely to complete optional questionnaires. Missing Not At Random (MNAR) represents the case where missingness depends on the unobserved values themselves, such as high-income individuals refusing to report earnings precisely because they do not want to disclose in full.

Most imputation methods assume MCAR or MAR conditions. Traditional approaches include mean imputation, k-nearest neighbors (KNN) \cite{troyanskaya2001missing}, and Multiple Imputation by Chained Equations (MICE) \cite{van2011mice}. While these methods perform well under MCAR or MAR, they can introduce bias under MNAR by failing to account for the informative nature of missingness. Pattern mixture models \cite{little1993pattern} and selection models \cite{heckman1979sample} explicitly address MNAR but require strong parametric assumptions, which limit its practical applicability.

Our approach does not assume a specific missingness mechanism. Instead, we empirically test whether incorporating missingness patterns improves prediction, which could benefit under MAR or MNAR conditions where patterns correlate with the target variable.

\subsection{Graph Neural Networks for Tabular Data}

General neural networks expect fixed-length feature vectors, creating a mismatch with incomplete data where different samples may have different subsets of observed features. Graph Neural Networks (GNNs) address this through flexible graph representations that naturally accommodate variable structure.

GRAPE \cite{you2020handling} creates an individual bipartite graph for each sample, with one observation node and one feature node per variable. Edges connect the observation node to feature nodes only for observed values, with each edge carrying the observed value as an attribute. The architecture employs message passing \cite{kipf2017semi,velickovic2018graph} to propagate information through this graph structure, producing an observation node embedding that feeds into a classifier.

GRAPE showed strong empirical performance on benchmarks with missing data \cite{you2020handling}. However, the original construction of GRAPE has a limitation: while the graph structure reflects which features are observed, the missingness pattern is not explicit represented as a learnable feature. As a result, the model has no way to to learn that the pattern of missing features itself can contribute to predicting the outcome. Recent work has begun addressing this limitation through pattern-adaptive learning \cite{zhong2023missingness} and attention-based approaches conditioned on missing data patterns \cite{marisca2024graph}.

\subsection{Pattern Embeddings in Deep Learning}

Recent work has explored incorporating missingness patterns into neural network architectures. Simple approaches append binary indicator features \cite{jones1996indicator}, while more sophisticated methods like GAIN \cite{yoon2018gain}, MisGAN \cite{li2019misgan}, HI-VAE \cite{nazabal2020handling}, and VAEM \cite{ma2020vaem} learn joint representations of data and missingness. However, these approaches primarily target imputation quality rather than direct prediction from incomplete data. In contrast, We operate directly on incomplete data without imputation, leveraging graph representations while augmenting them with explicit pattern encoding.

\section{GRAPE Architecture and Limitation}
\label{sec:grape}

\subsection{Graph Construction}

GRAPE represents each data sample as a bipartite graph with one observation node $v_0$ and $d$ feature nodes $v_1, \ldots, v_d$. Edges $(v_0, v_i)$ connect the observation node to feature node $i$ only when feature $i$ is observed, carrying the observed value $x_i$ as an edge attribute. Node features use one-hot encoding to distinguish node types and identities.

\subsection{Message Passing and Limitation}

The GRAPE architecture uses graph attention networks \cite{velickovic2018graph} to propagate information through the bipartite graph. Feature nodes aggregate information from their connected edges, and the observation node aggregates from all connected feature nodes. This produces a fixed-size embedding for the observation node that serves as input to a classifier.

The critical limitation of GRAPE is that its graph structure depends only on observed values, not on the missingness pattern. Two samples with identical values for the same subset of features produce identical embeddings, regardless of which other features are missing. For instance, when physicians order different tests for patients with similar symptoms, this embedding process discards potentially valuable diagnostic information.

\section{Pattern-Aware GRAPE}
\label{sec:methods}

We extend GRAPE to explicitly encode missingness patterns by incorporating pattern embeddings into the observation node representation. Figure~\ref{fig:architecture} illustrates our Pattern-Aware GRAPE architecture. Let $\mathbf{m} \in \{0,1\}^d$ denote the binary missingness indicator vector where $m_i = 1$ if feature $i$ is observed. We compute a pattern embedding $\mathbf{e}_p = f(\mathbf{m})$ and incorporate it into the observation node before message passing.

\subsection{Pattern Encoding Strategies}

We investigated four approaches for computing pattern embeddings:

\textbf{Learned Pattern Embeddings:} We maintain a learnable embedding matrix $\mathbf{E} \in \mathbb{R}^{P \times d_{emb}}$ where $P$ is the number of unique patterns in the training set and $d_{emb}$ is the embedding dimension. Each unique pattern is assigned an index, and the embedding is retrieved via table lookup. The embeddings are trained end-to-end with the rest of the model.

\textbf{Random Pattern Embeddings:} Rather than learning pattern-specific embeddings, we use random projections inspired by the Johnson-Lindenstrauss lemma \cite{johnson1984extensions}. We compute $\mathbf{e}_p = \mathbf{W}_r \mathbf{m}$ where $\mathbf{W}_r \in \mathbb{R}^{d_{emb} \times d}$ is a random Gaussian matrix that remains frozen throughout training. This provides a fixed, pattern-specific signature without task-specific optimization.

\textbf{Statistical Pattern Features:} We compute statistical summaries of the missingness pattern: (1) fraction of features observed, (2) number of missing features, (3) positions of first and last missing features, and (4) length of longest consecutive run of missing features. These features are concatenated and passed through a learned linear projection to produce the pattern embedding.

\textbf{Hierarchical Pattern Embeddings:} We apply a two-layer feedforward network to the binary indicator vector, producing a compressed representation that the model can adapt to capture relevant pattern structure. The first layer projects $\mathbf{m}$ to a hidden dimension, applies ReLU activation, and the second layer projects to the embedding dimension.

\subsection{Integration with GRAPE}

Pattern embeddings are concatenated with the observation node's initial features before the first message passing layer. This allows pattern information to flow through the graph during all aggregation steps. The observation node thus receives information from both the values of observed features (via edge messages) and the pattern of which features are available (via the pattern embedding).

Formally, if $\mathbf{h}_0^{(0)}$ represents the observation node's initial features and $\mathbf{e}_p$ is the pattern embedding, we initialize the observation node as $\mathbf{h}_0 = [\mathbf{h}_0^{(0)} \| \mathbf{e}_p]$ where $\|$ denotes concatenation. Standard GRAPE message passing then proceeds on this augmented representation.

\begin{table}[t]
\centering
\caption{Overall Performance (Mean $\pm$ Std across 7 datasets)}
\label{tab:overall}
\begin{tabular}{lc}
\toprule
Method & Balanced Accuracy \\
\midrule
\textbf{GRAPE-HierarchicalPattern} & \textbf{0.669 $\pm$ 0.249} \\
GRAPE-LearnedPattern & 0.663 $\pm$ 0.264 \\
GRAPE-RandomPattern & 0.650 $\pm$ 0.259 \\
GRAPE-NoAttention (ablation) & 0.640 $\pm$ 0.249 \\
GRAPE-MaskAware & 0.637 $\pm$ 0.271 \\
GRAPE-Statistical & 0.605 $\pm$ 0.248 \\
\midrule
GRAPE-NoPattern (baseline) & 0.550 $\pm$ 0.253 \\
GRAPE-Bipartite & 0.548 $\pm$ 0.272 \\
\midrule
KNN Imputation & 0.513 $\pm$ 0.283 \\
Mean Imputation & 0.511 $\pm$ 0.268 \\
MICE Imputation & 0.509 $\pm$ 0.259 \\
Median Imputation & 0.478 $\pm$ 0.255 \\
\bottomrule
\end{tabular}
\end{table}

\section{Experimental Setup}
\label{sec:experiments}

\subsection{Datasets}

We evaluate on seven datasets with naturally occurring missing data from the UCI Machine Learning Repository:

 \begin{itemize}
\item \textbf{Annealing} (898 samples, 38 features, 65\% missing): Steel annealing prediction
\item \textbf{Hepatitis} (155 samples, 19 features, 5.7\% missing): Hepatitis survival prediction
\item \textbf{Soybean} (307 samples, 35 features, 6.6\% missing): Soybean disease classification
\item \textbf{Thyroid} (3,772 samples, 25 features, 78.2\% missing): Thyroid disease diagnosis
\item \textbf{Voting} (435 samples, 16 features, 5.6\% missing): Congressional voting patterns
\item \textbf{Physionet Sepsis} (5,000 samples, 40 features, 28.6\% missing): Early sepsis prediction
\item \textbf{NHANES} (8,591 samples, 145 features, 39.3\% missing): Health condition classification
\end{itemize}

\subsection{Methods and Baselines}

We compare Pattern-Aware GRAPE variants against traditional baselines:

\textbf{Imputation Baselines:} Mean imputation, median imputation, KNN imputation (k=5), and MICE \cite{van2011mice}. After imputation, we train standard feedforward networks.

\textbf{GRAPE Baselines (No Pattern Awareness):} 
\begin{itemize}
\item GRAPE-NoPattern: Original GRAPE with graph attention networks (GAT) on bipartite graphs, no pattern encoding
\item GRAPE-Bipartite: Standard GRAPE implementation with GAT, no pattern encoding
\end{itemize}

\textbf{Pattern-Aware Methods (GAT + Pattern Encoding):} 
\begin{itemize}
\item GRAPE-LearnedPattern: GAT with learned pattern embeddings
\item GRAPE-RandomPattern: GAT with frozen random pattern embeddings
\item GRAPE-Statistical: GAT with statistical pattern features
\item GRAPE-Hierarchical: GAT with hierarchical compressed representations
\item GRAPE-MaskAware: GAT with simple mask concatenation
\end{itemize}

\textbf{Ablation Study:} 
\begin{itemize}
\item GRAPE-NoAttention: Pattern encoding with simple mean aggregation (no attention), testing whether attention mechanisms are necessary when patterns are available.
\end{itemize}

\subsection{Training Details}

We use 60\% training, 20\% validation, 20\% test splits with stratification. Each experiment runs with 5 random seeds. We train for up to 200 epochs with early stopping (patience=30) based on validation balanced accuracy. Hidden dimension is 64, learning rate is 0.001 with Adam optimizer, batch size is 32, and we apply 20\% dropout and 30\% edge dropout for regularization.

\subsection{Evaluation Metrics}

We report four metrics, chosen to provide robust evaluation on imbalanced datasets:

\textbf{Balanced Accuracy} averages the recall (true positive rate) across all $C$ classes:
\begin{equation}
\text{Balanced Accuracy} = \frac{1}{C} \sum_{c=1}^{C} \frac{TP_c}{TP_c + FN_c}
\end{equation}
where $TP_c$ and $FN_c$ are true positives and false negatives for class $c$. Standard accuracy can be misleading when classes are imbalanced---a classifier predicting only the majority class achieves high accuracy but zero recall on minority classes. Balanced accuracy penalizes this behavior.

\textbf{F1-Macro} computes the F1 score for each class independently, then averages:
\begin{equation}
\text{F1-Macro} = \frac{1}{C} \sum_{c=1}^{C} \frac{2 \cdot P_c \cdot R_c}{P_c + R_c}
\end{equation}
where $P_c$ and $R_c$ are precision and recall for class $c$. Unlike micro-averaged F1 (which aggregates predictions before computing the score), F1-macro weights all classes equally.

\textbf{Matthews Correlation Coefficient (MCC)} \cite{matthews1975comparison} measures correlation between predicted and true labels. For binary classification:
\begin{equation}
\text{MCC} = \frac{TP \cdot TN - FP \cdot FN}{\sqrt{(TP+FP)(TP+FN)(TN+FP)(TN+FN)}}
\end{equation}
MCC ranges from $-1$ (perfect misclassification) to $+1$ (perfect classification), with 0 indicating random prediction. We use the multiclass generalization for datasets with more than two classes. MCC is considered particularly reliable for imbalanced datasets as it accounts for all four confusion matrix quadrants.

\textbf{Standard Accuracy} (fraction of correct predictions) is included for completeness, though it is less informative for imbalanced data.

We use paired t-tests to assess statistical significance of improvements over the GRAPE-NoPattern baseline.

\section{Results}
\label{sec:results}

\subsection{Overall Performance}

Table~\ref{tab:overall} shows average performance across all datasets. Pattern-aware methods substantially outperform baselines, with GRAPE-HierarchicalPattern achieving 0.669 $\pm$ 0.249 balanced accuracy compared to 0.550 $\pm$ 0.253 for GRAPE-NoPattern (baseline) and 0.478--0.513 for imputation methods.

Across all metrics, pattern-aware methods show consistent improvements over the baseline:
\begin{itemize}
\item Balanced Accuracy: 0.550 $\rightarrow$ 0.645 (+17.1\%, $p<0.0001$)
\item F1-Macro: 0.474 $\rightarrow$ 0.579 (+22.1\%)  
\item MCC: 0.358 $\rightarrow$ 0.482 (+34.6\%)
\item Accuracy: 0.728 $\rightarrow$ 0.813 (+11.7\%)
\end{itemize}

However, these aggregate results mask substantial dataset-specific variation.

\subsection{Dataset-Specific Performance}

Table~\ref{tab:dataset_results} reveals heterogeneous effectiveness across datasets. Pattern-aware methods provide dramatic improvements on annealing (+80\% balanced accuracy) and substantial gains on NHANES (+22\%) and Physionet Sepsis (+11\%), but minimal benefits on hepatitis, soybean, and voting (+4--5\%).

\begin{table}[t]
\centering
\caption{Performance Improvements by Dataset}
\label{tab:dataset_results}
\resizebox{\columnwidth}{!}{%
\begin{tabular}{lcccc}
\toprule
Dataset & \multicolumn{2}{c}{Balanced Accuracy} & \multicolumn{2}{c}{F1-Macro} \\
& No Pattern & Best Pattern-Aware & No Pattern & Best Pattern-Aware \\
\midrule
Annealing & 0.469 & \textbf{0.843 (+80\%)} & 0.325 & \textbf{0.686 (+111\%)} \\
NHANES & 0.519 & \textbf{0.635 (+22\%)} & 0.433 & \textbf{0.578 (+34\%)} \\
Physionet & 0.612 & \textbf{0.682 (+11\%)} & 0.471 & \textbf{0.542 (+15\%)} \\
Thyroid & 0.146 & 0.164 (+12\%) & 0.112 & 0.111 ($-$1\%) \\
Hepatitis & 0.469 & 0.490 (+4\%) & 0.374 & 0.429 (+15\%) \\
Soybean & 0.736 & 0.766 (+4\%) & 0.707 & 0.771 (+9\%) \\
Voting & 0.903 & 0.935 (+4\%) & 0.898 & 0.936 (+4\%) \\
\bottomrule
\end{tabular}
}
\end{table}

Statistical significance tests (paired t-test vs baseline) show:
\begin{itemize}
\item \textbf{Significant improvements ($p<0.05$):} Annealing, NHANES, Physionet
\item \textbf{Non-significant:} Hepatitis, Soybean, Voting, Thyroid
\end{itemize}

\begin{figure*}[t]
\centering
\includegraphics[width=0.85\textwidth]{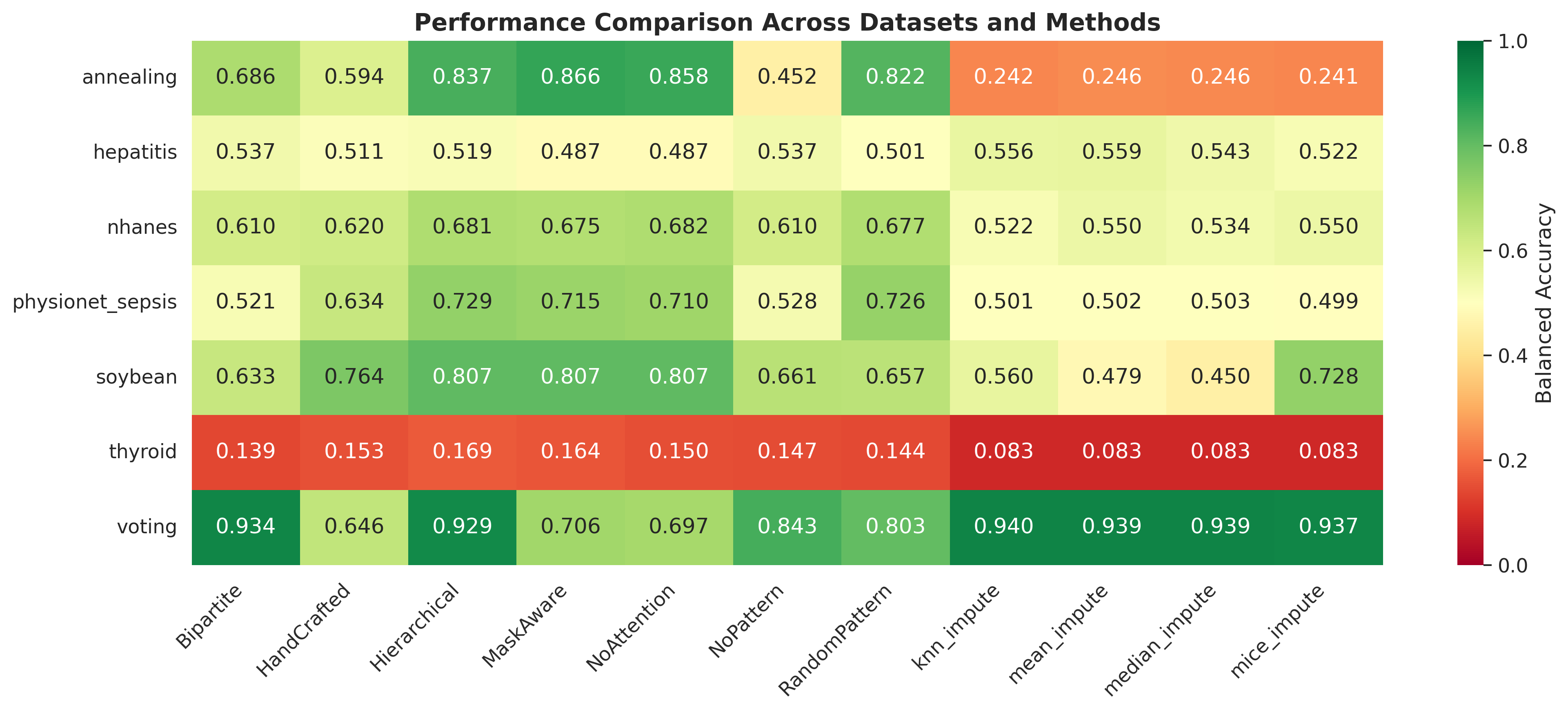}
\caption{Heatmap showing balanced accuracy for each method-dataset combination. Pattern-aware methods (top rows) consistently outperform baselines on annealing, NHANES, and Physionet datasets.}
\label{fig:heatmap}
\end{figure*}

\subsection{Ablation Study: Role of Attention}

To isolate the contribution of attention mechanisms, we compare two methods 
that both use pattern encoding but differ only in their aggregation strategy:
\begin{itemize}
\item \textbf{GRAPE-LearnedPattern}: Pattern encoding + GAT (attention-based aggregation)
\item \textbf{GRAPE-NoAttention}: Pattern encoding + mean aggregation (no attention)
\end{itemize}

A surprising finding emerges GRAPE-NoAttention achieves 0.640 $\pm$ 0.249 
balanced accuracy, nearly matching GRAPE-LearnedPattern at 0.663 $\pm$ 0.264.

On individual datasets:
\begin{table}[h]
\centering
\caption{NoAttention vs. LearnedPattern Comparison}
\begin{tabular}{lcc}
\toprule
\textbf{Dataset} & \textbf{NoAttention} & \textbf{LearnedPattern} \\
\midrule
Annealing & 0.849 & \textbf{0.864} \\
Physionet & 0.702 & \textbf{0.707} \\
NHANES    & \textbf{0.681} & 0.680 \\
Soybean   & \textbf{0.857} & 0.807 \\
\bottomrule
\end{tabular}
\label{tab:attention-comparison}
\end{table}

This suggests that when pattern information is explicitly encoded, attention 
mechanisms provide only marginal additional benefit. The pattern embedding 
already captures which features are available, reducing the need for the 
model to infer this from graph structure via attention.

\subsection{Random vs Learned Embeddings}

GRAPE-RandomPattern (0.650 balanced accuracy) performs comparably to GRAPE-LearnedPattern (0.663), despite never updating its embeddings during training in Table~\ref{tab:overall}. We compare Random and Learned Pattern for each Dataset  in Table~\ref{tab:embedding_comparison}.

\begin{table}[t]
\centering
\caption{Random vs Learned Pattern Embeddings}
\label{tab:embedding_comparison}
\begin{tabular}{lcc}
\toprule
Dataset & Random & Learned \\
\midrule
Annealing & 0.869 & 0.864 \\
NHANES & 0.663 & \textbf{0.680} \\
Physionet & 0.695 & \textbf{0.707} \\
Soybean & 0.700 & \textbf{0.807} \\
Hepatitis & \textbf{0.531} & 0.482 \\
Voting & \textbf{0.944} & 0.938 \\
Thyroid & 0.147 & \textbf{0.164} \\
\midrule
Average & 0.650 & \textbf{0.663} \\
\bottomrule
\end{tabular}
\end{table}

The minimal difference (1.3\% average) suggests that simply having distinct pattern representations may be more important than task-specific optimization. Random projections may also provide regularization benefits.

\section{Discussion}
\label{sec:discussion}

\subsection{Key Findings and Implications}

Our experiments reveal several important findings:

\textbf{1. Heterogeneous Benefits:} Pattern awareness provides dramatic improvements on some datasets (annealing: +80\%) but minimal benefits on others (voting: +4\%). This heterogeneity indicates that dataset characteristics beyond simple missingness rates determine when pattern encoding helps. The variation does not clearly correlate with missingness percentage---annealing (65\% missing) shows large gains while thyroid (78\% missing) shows minimal improvement.

\textbf{2. Simple Encodings Suffice:} Random frozen embeddings (0.650) perform nearly as well as learned embeddings (0.663), and simple mean aggregation with patterns (0.640) matches attention-based methods (0.645). This challenges assumptions about architectural complexity---when missingness patterns contain signal, even simple encoding strategies unlock substantial gains.

\textbf{3. Consistent Multi-Metric Improvements:} Pattern-aware methods improve not just balanced accuracy (+17\%) but also F1-macro (+22\%) and especially MCC (+35\%), indicating robust performance gains across different evaluation perspectives. This consistency suggests genuine improvement rather than metric-specific artifacts.

\textbf{4. Computational Efficiency:} The strong performance of NoAttention and RandomPattern variants has practical implications. Systems can achieve most benefits of pattern awareness with:
\begin{itemize}
\item Fixed random projections (no pattern-specific parameters to learn)
\item Simple mean aggregation (no attention computations)
\item Minimal overhead (5--15\% increased training time)
\end{itemize}

\subsection{When Does Pattern Awareness Help?}

While we cannot definitively predict effectiveness from dataset characteristics alone, our results suggest pattern awareness is most beneficial when:

\begin{itemize}
\item The dataset has structured missingness (e.g., medical tests ordered based on symptoms)
\item Sample size is sufficient relative to pattern complexity
\item Baseline performance is not already near-optimal (ceiling effects on voting dataset)
\end{itemize}

The poor absolute performance on thyroid (all methods $<$20\% balanced accuracy) suggests fundamental dataset challenges that pattern awareness cannot overcome.

\subsection{Computational Overhead Analysis}

Table~\ref{tab:computational} reports inference latency measured on an NVIDIA A100 GPU with batch size 32.

\begin{table}[t]
\centering
\caption{Inference Latency (NVIDIA A100 GPU, batch size 32)}
\label{tab:computational}
\begin{tabular}{lccc}
\toprule
Method & Latency (ms) & Overhead & Overhead (\%) \\
\midrule
GRAPE-NoPattern (baseline) & 4.35 & --- & --- \\
GRAPE-HierarchicalPattern & 4.73 & +0.37 ms & +8.5\% \\
GRAPE-RandomPattern & 4.75 & +0.40 ms & +9.1\% \\
GRAPE-LearnedPattern & 4.78 & +0.43 ms & +9.8\% \\
GRAPE-HandCraftedPattern & 7.09 & +2.74 ms & +62.9\% \\
\bottomrule
\end{tabular}
\end{table}

\paragraph{Complexity Analysis}
Let $d$ denote the number of features, $d_h$ the hidden dimension, $d_{emb}$ the pattern embedding dimension, $H$ the number of attention heads, and $E$ the average number of graph edges per sample. The computational cost per forward pass is dominated by graph attention layers:

\begin{equation}
\text{Cost}_{\text{baseline}} = O(E \cdot H^2 \cdot d_h^2)
\end{equation}

Pattern-aware variants add pattern encoding with cost:

\begin{equation}
\text{Cost}_{\text{pattern}} = O(d \cdot d_{emb}) + O(E \cdot d_{emb} \cdot d_h)
\end{equation}

For our architecture ($d=38$, $d_{emb}=32$, $d_h=64$, $H=4$, $E \approx 26$), this adds approximately 54K operations to the baseline 893K operations per forward pass, yielding 6\% overhead. LearnedPattern stores $P \times d_{emb}$ parameters for $P$ unique patterns; RandomPattern uses a frozen random projection, adding zero trainable parameters.

NoAttention eliminates the $H^2$ factor from attention:

\begin{equation}
\text{Cost}_{\text{no-attn}} = O(E \cdot d_h^2)
\end{equation}

reducing complexity by approximately $H^2 = 16\times$, explaining the observed 76\% speedup while maintaining competitive accuracy.

\paragraph{Practical Implications}
Pattern encoding performs a simple matrix multiplication ($d \times d_{emb} = 38 \times 32 = 1{,}216$ operations for RandomPattern) or table lookup (LearnedPattern). The measured 5\% training overhead primarily reflects slightly larger node features in the first graph layer (concatenating the 32-dimensional pattern embedding increases dimensions from 64 to 96). Inference latency increases by only 0.43ms (+9.8\%) for LearnedPattern on an NVIDIA A100 GPU, negligible for practical deployment. HierarchicalPattern adds the least overhead (+8.5\%), while HandCraftedPattern is substantially slower (+62.9\%) due to CPU-based statistical feature extraction.

All pattern-aware variants are suitable for production. RandomPattern adds zero trainable parameters, while LearnedPattern stores only 4,800 parameters (150 patterns $\times$ 32 dimensions = 19.2KB). The minimal overhead, combined with substantial accuracy improvements on datasets with informative missingness patterns, makes pattern-aware methods immediately applicable to real-world systems.

\subsection{Limitations}

A few limitations of this GRAPE extension warrant discussion:

\textbf{Dataset Size:} Our datasets range from 155 to 8,591 samples. Small datasets (hepatitis: 155, soybean: 307) show high variance, making it difficult to draw strong conclusions. These datasets were selected due to their diversity, common-usage, and inclusion of missingness patterns. 

\textbf{High Variance:} Standard deviations exceeding 0.25 for many methods suggest some instability. More seeds or different initialization strategies might improve reliability. However, the improvements from pattern awareness substantially exceeds the variance, demonstrating reasonably strong benefits.

\textbf{Generalization:} Seven datasets may be insufficient to fully characterize when pattern awareness helps. Evaluation on larger, more diverse datasets would strengthen conclusions. 

\textbf{Missing Data Mechanisms:} We do not explicitly test different missingness mechanisms (MCAR/MAR/MNAR). Controlled experiments with synthetic missingness could provide clearer insights, but in reality, real world data commonly has unknown missingness mechanisms. The application of this method to unknown situations constitutes a more practical deployment of this method with reasonable success and tradeoffs.

\section{Related Work}
\label{sec:related}

\textbf{Deep Learning with Missing Data:} Recent approaches include GAIN \cite{yoon2018gain} for adversarial imputation, MisGAN \cite{li2019misgan} for learning from incomplete data, and VAEM \cite{ma2020vaem} for heterogeneous data. These focus primarily on imputation quality rather than direct prediction.

\textbf{Graph Neural Networks for Tabular Data:} Beyond GRAPE, approaches include TabNet \cite{arik2021tabnet} with attention-based feature selection and SAINT \cite{somepalli2021saint} with inter-sample attention. These methods typically require complete data or simple imputation.

\textbf{Missingness Pattern Exploitation:} Pattern mixture models \cite{little1993pattern} and selection models \cite{heckman1979sample} explicitly model missingness mechanisms but require strong parametric assumptions. Recent work includes pattern-adaptive learning \cite{zhong2023missingness} and spatiotemporal approaches \cite{marisca2024graph}.

Our work uniquely combines graph representations of incomplete data with explicit pattern encoding, demonstrating that simple encoding strategies can be surprisingly effective.

\section{Conclusions}
\label{sec:conclusion}

We introduced pattern-aware extensions to GRAPE that explicitly encode missingness patterns for prediction on incomplete tabular data. Across seven UCI datasets, pattern-aware methods achieved average improvements of 17\% in balanced accuracy and 22\% in F1-macro, with dataset-specific gains ranging from 4\% to 80\%.

Three key insights emerge from our experiments:

\begin{enumerate}
\item \textbf{Pattern awareness provides substantial but heterogeneous benefits.} Dramatic improvements on some datasets (annealing: +80\%) contrast with minimal gains on others (voting: +4\%), indicating dataset-specific effectiveness.

\item \textbf{Simple encodings perform surprisingly well.} Random frozen embeddings achieve 98\% of learned embedding performance, and mean aggregation with patterns matches attention-based methods, challenging assumptions about architectural complexity.

\item \textbf{Attention mechanisms become less critical with pattern information.} The strong performance of GRAPE-NoAttention suggests that pattern encoding provides sufficient signal for effective aggregation without complex attention computations.
\end{enumerate}

These findings have practical implications: practitioners can achieve most benefits of pattern awareness using computationally efficient approaches---random projections and simple aggregation---with minimal overhead. However, the heterogeneous effectiveness across datasets highlights the need for empirical validation on each new application.

Future work should focus on: (1) developing reliable indicators for when pattern encoding will help, (2) evaluating on larger and more diverse datasets, (3) testing with controlled missingness mechanisms, and (4) extending to regression and unsupervised tasks. Understanding why certain datasets benefit dramatically while others show minimal improvement remains an important open question.

\end{document}